# Real-Time Bundle Adjustment for Ultra-High-Resolution UAV Imagery Using Adaptive Patch-Based Feature Tracking

Selim Ahmet Iz[1,2], Francesco Nex[2], Norman Kerle[2], Henry Meissner[1], Ralf Berger[1]

[1]German Aerospace Center (DLR), Institute of Optical Sensor Systems, Berlin, Germany
(selim.iz, henry.meissner, ralf.berger)@dlr.de
[2]Faculty of Geo-Information Science and Earth Observation (ITC), University of Twente, Enschede, The Netherlands
(f.nex, n.kerle)@utwente.nl

**Abstract**

Real-time processing of UAV imagery is crucial for applications requiring urgent geospatial information, such as disaster response, where rapid decision-making and accurate spatial data are essential. However, processing high-resolution imagery in real time presents significant challenges due to the computational demands of feature extraction, matching, and bundle adjustment (BA). Conventional BA methods either downsample images, sacrificing important details, or require extensive processing time, making them unsuitable for time-critical missions. To overcome these limitations, we propose a novel real-time BA framework that operates directly on full-resolution UAV imagery without downsampling. Our lightweight, onboard-compatible approach divides each image into user-defined patches (e.g., NxN grids, default 150×150 pixels) and dynamically tracks them across frames using UAV GNSS/IMU data and a coarse, globally available digital surface model (DSM). This ensures spatial consistency for robust feature extraction and matching between patches. Overlapping relationships between images are determined in real time using UAV navigation system, enabling the rapid selection of relevant neighbouring images for localized BA. By limiting optimization to a sliding cluster of overlapping images, including those from adjacent flight strips, the method achieves real-time performance while preserving the accuracy of global BA. The proposed algorithm is designed for seamless integration into the DLR Modular Aerial Camera System (MACS), supporting large-area mapping in real time for disaster response, infrastructure monitoring, and coastal protection. Validation on MACS datasets with 50MP images demonstrates that the method maintains precise camera orientations and high-fidelity mapping across multiple strips, running full bundle adjustment in under 2 seconds without GPU acceleration.

## 1. Introduction

The demand for real-time aerial mapping and rapid 3D reconstruction is steadily increasing in various fields such as disaster management, infrastructure monitoring, agriculture, and coastal protection. In time-critical scenarios such as disaster response, timely and accurate geospatial data are essential for informed decision-making under dynamic and uncertain conditions. Fast generation of georeferenced aerial imagery and 3D models helps first responders localize damages, plan rescue operations, and assess infrastructure stability in near real-time. While emergency response is a primary use case, rapid mapping is also beneficial for routine applications where timely spatial insights can improve operational efficiency. One of the most critical components in real-time mapping workflows is the estimation of accurate image orientations during flight, which is the key prerequisite for generating consistent 3D point clouds and georeferenced orthophotos.

In these applications, real-time refers to the ability to process all captured images during the flight, a requirement that poses substantial computational challenges. These challenges stem from variability in flight and camera parameters such as image size, flight velocity, image acquisition frequency and flight altitude (overlap ratio) as can be seen in Eq. (1) and (2). In a representative scenario with a flight velocity of 20 m/s, flight altitude of 300 m, and 80% overlap in the flight direction, the maximum target processing time per image pair can be computed as 2.08 seconds.

$$t = h_{im}(1 - \beta/100)/v \quad (1)$$

where t is the processing time (sec), $h_{im}$ is the image footprint in flight direction (m), $v$ is velocity (m/s) and $\beta$ is overlap ratio (%) which can be derived as:

$$\beta_\% = \left(1 - \frac{10 f_{mm} v \Delta t}{N \theta_{\mu m} h_{flight}}\right) x 100 \quad (2)$$

where $f_{mm}$ is the focal length (mm), $\Delta t$ is the camera shooting frequency (s), $N$ is the number of pixels along flight direction, $\theta_{\mu m}$ is the pixel size ($\mu m$) and finally $h_{flight}$ is the flight altitude (m).

To address these needs, the Institute of Optical Sensor Systems at the German Aerospace Center (DLR) has developed modular aerial imaging systems and workflows (MACS-Mosaica) that support rapid and accurate georeferencing of UAV imagery (Hein & Berger, 2018). These systems are designed to operate with lightweight onboard hardware and focus on minimizing the latency between image acquisition and map generation. Notably, recent developments such as terrain-aware image clipping (TAC) enable real-time map generation by geometrically intersecting individual images with a DSM to extract the most relevant rectangular sections of each frame. This method requires no bundle adjustment and provides high accuracy within individual flight strips, offering a practical solution for scenarios where immediate image transmission or post-landing map generation is necessary (Figure 1, except step 4). However, it does not address the need for consistent orientation estimation between overlapping images across different strips, a crucial requirement not only for improving users' comprehensive understanding of the 2D scene, but also for accurate 3D mapping.

Conventional BA algorithms, while robust for offline workflows, struggle to meet stringent real-time constraints. For example, running full-resolution global BA on a typical acquisition of high-resolution (e.g. MACS images at around 50MP) images can take several minutes to converge, far exceeding our target of approximately 2 seconds for near-instantaneous map updates.

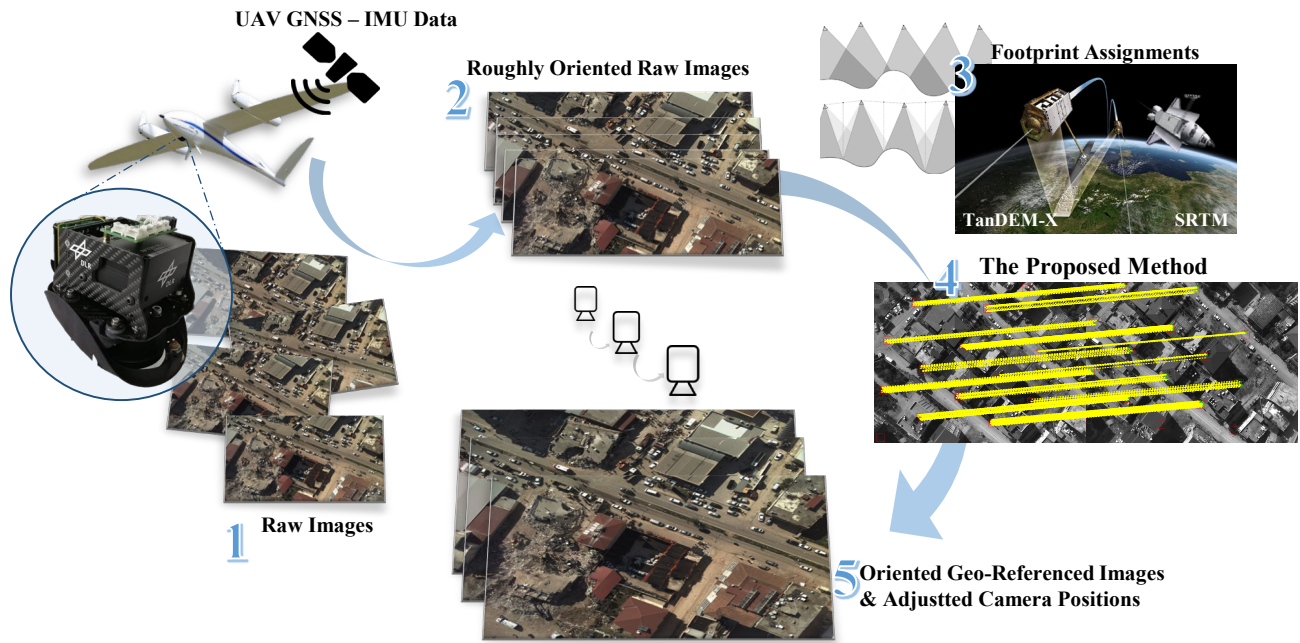

Figure 1: Workflow overview

This is due to the dense optimization over all available camera poses and 3D points, which quickly becomes computationally prohibitive. While downsampling strategies can reduce computation time, they introduce a trade-off by degrading spatial precision, an unacceptable compromise for tasks requiring high-detail surface analysis or accuracy (d'Angelo & Kurz, 2019).

To overcome these limitations, we present a novel local bundle adjustment framework that achieves real-time processing without downsampling, preserving the full resolution and detail of the original imagery (step 4 in Figure 1). Our method introduces a patch-based approach that divides each image into an adaptive NxN grid of patches, which are individually tracked across sequential frames. These patches are spatially guided using UAV GNSS/IMU measurements and a DSM, ensuring that the same ground regions are followed through the image sequences. Rather than relying solely on direct navigation data (Iz & Munel, 2023), we incorporate image footprint information, specifically, the projection of corner world coordinates onto an a-priori known elevation model (e.g., TanDEM-X or SRTM-Shuttle Radar Topography Mission), to compute inter-image transformations more accurately. This geo-referenced strategy enhances spatial consistency in patch tracking, and significantly improves feature matching performance, particularly between strips.

The proposed framework offers a balanced solution and contributes to the literature on real-time bundle adjustment with high-resolution images in the following aspects. It

- introduces a novel localized bundle adjustment strategy that operates on sliding clusters of overlapping images, including across adjacent flight strips when applicable, during flight.
- provides geo-referenced multi-strip image orientation solution with translated patches using each image's ECEF (Earth-Centered Earth-Fixed) corner coordinates.
- accelerates matching, to strike an optimal accuracy–efficiency balance.
- provides an adaptable solution for varying UAV image resolutions through user-defined patch parameters, allowing flexibility in deployment scenarios with high-resolution sensors.
- supports seamless integration into existing systems, such as the DLR's MACS (Lehmann et al., 2011), enabling applicability in operational large-scale mapping tasks.

The remainder of this paper is organized as follows: Section II reviews related work on bundle adjustment and real-time photogrammetry. Section III describes the proposed patch-based tracking and localized BA method. Section IV presents experimental results and performance evaluation. Finally, Section V concludes the paper and suggests directions for future work.

## 2. Related Work

Real-time mapping from UAVs has long been a subject of significant interest across photogrammetry, robotics, and remote sensing. Traditional mapping systems have largely relied on offline processing pipelines, often requiring the entire dataset to be available before processing can begin. These pipelines, including incremental and global Structure-from-Motion (SfM) methods, have achieved impressive results in reconstructing high-fidelity 3D models from UAV imagery (Snavely et al., 2006; Wu, 2011). However, their dependence on full dataset availability and computationally intensive BA steps makes them unsuitable for dynamic, real-time missions.

Efforts to migrate from offline SfM to real-time pipelines have leveraged visual SLAM frameworks that approximate camera poses incrementally using onboard sensors. Notably, several SLAM-based systems, such as ORB-SLAM variants and Map2DFusion, have demonstrated their ability to perform image stitching and sparse reconstruction on low-resolution aerial sequences (Mur-Artal & Tardós, 2017; Bu et al., 2016). These systems are particularly effective in high-overlap image sequences captured at lower altitudes. Nevertheless, they often suffer from accumulated drift and poor performance under low-texture or high-parallax conditions (Zhao et al., 2023).

To enhance robustness, many mapping frameworks integrate auxiliary sensors such as GNSS and IMUs into the SLAM pipeline. This coupling mitigates localization errors and improves consistency of global map alignment (Hinzmann et al., 2016; Wu et al., 2020). Yet, even hybrid systems struggle when real-time constraints are combined with high-resolution imaging requirements. Most current approaches, including those that exploit GPU acceleration or matching confidence in dense feature tracking, still require significant computational resources (Zhao et al., 2023; Yao et al., 2019). This makes them impractical for onboard processing, especially when high-resolution images beyond 50 megapixels are involved.

Other approaches have attempted to bypass heavy photogrammetric processing by simplifying alignment to 2D homography-based mosaicking. These systems achieve rapid visual output, but often compromise on geometric accuracy and cannot handle significant parallax or topographic variation (Ghosh & Kaabouch, 2016; Kekec et al., 2014). A notable exception is the TAC workflow developed by Hein and Berger (2019), which offers a topography-aware, geo-referenced image processing method suitable for onboard execution. By intersecting image footprints with a digital terrain model, the TAC method enables fast and accurate map generation without the need for bundle adjustment. Nevertheless, as noted earlier, it does not incorporate inter-strip feature matching or global orientation estimation, which limits its effectiveness in achieving coherent alignment across overlapping strips, an essential requirement for seamless 2D scene understanding and 3D reconstruction.

A further frontier lies in the application of dense stereo and depth-enhanced mapping pipelines. SLAM-integrated depth processing, as seen in TerrainFusion and OpenREALM, enables live 3D mesh generation or point cloud fusion, though often at the cost of reduced resolution or delayed processing (Wang et al., 2016; Kern et al., 2020). While effective in moderate-resolution cases, these systems still rely on significant onboard computational resources, frequently requiring dedicated GPUs, post-flight data refinement, data-driven models, or fast rendering techniques.

More recent 3D reconstruction pipelines based on Neural Radiance Fields (NeRFs) have gained traction for their ability to reconstruct detailed scenes from sparse inputs. Systems such as FlyNeRF (Dronova et al., 2024) and UAV-NeRF (Li et al., 2024b) integrate flight path planning and incident-angle-aware sampling to improve rendering quality and depth estimation from drone-captured imagery. However, these methods primarily target photorealistic rendering and DSM generation, relying on iterative volumetric optimization, GPU acceleration, and pre-oriented image sets; typically using lower-resolution images than those considered in our work.

Similarly, Gaussian Splatting has rapidly evolved as a lightweight yet highly photorealistic alternative for scene modeling and dense map generation. DroneSplat (Tang et al., 2025) applies 3D Gaussian splatting in multi-view aerial settings, guided by stereo priors and visibility prediction. It handles limited baselines and dynamic distractors more robustly than NeRF in real-world UAV sequences. UAVTwin (Choi et al., 2025) further fuses splatting with synthetic human model injection for onboard perception tasks. Nonetheless, these approaches focus on novel-view rendering rather than real-time depth reconstruction, and still require inter-frame orientation information to initiate processing.

Finally, recent advances in feature matching have been driven by Transformer-based models and lightweight learned descriptors, enabling robust, dense correspondences in high-resolution UAV imagery. LoFTR is one of the touchstone studies to leverage attention mechanisms to handle large viewpoint and scale variations (Sun et al., 2021). More recent methods, such as LightGlue and XFeat, improve efficiency and modularity for large-scale aerial inputs, even exceeding 50MP, through decoupled architectures and multi-head attention (Lindenberger et al., 2023; Potje et al., 2024). These techniques, particularly when combined with hierarchical or patch-wise inference, strike a practical balance between accuracy and onboard feasibility, making them increasingly integral to modern aerial mapping pipelines. Historically, the idea of limiting tie point extraction to compact, well-distributed patterns also appeared in early photogrammetric approaches, such as Ebner's 3×3 configurations (Ebner, 1976) or Gruen's 5×5 grids (Gruen, 1985), even though our method was developed independently to address modern UAV-scale challenges.

## 3. Methods

In contrast to the existing body of work, our proposed method introduces a real-time, onboard-capable mapping pipeline specifically designed for ultra-high-resolution imagery (50MP+). Unlike prior systems constrained by fixed image sizes or GPU dependencies, our approach utilizes a user-defined dynamic patching mechanism that allows flexible and efficient memory allocation without compromising spatial resolution. This patch-wise image handling, combined with GNSS-aided direct geo-referencing, enables the generation of globally aligned outputs without the need for intensive global BA procedures.

Furthermore, our system performs continuous BA across overlapping image clusters during flight, accounting not only for current frames but also for adjacent strip images, thereby maintaining local consistency while enabling seamless integration into broader mapping frameworks. Crucially, the entire pipeline operates without dedicated GPU hardware, making it ideal for edge-based deployments and cost-sensitive UAV platforms. In the following subsections we detail the core components of our method, including patch-based feature extraction and tracking, footprint-aware spatial consistency, inter-strip integration, and final matching for BA.

### 3.1. Patch-Based Feature Extraction and Tracking

At the core of the proposed pipeline lies a flexible patch-based tracking strategy, in which the user defines a grid of patches over the initial image. Rather than processing the entire image globally, which becomes computationally infeasible for ultra-high-resolution imagery, the image is divided into a user-defined grid of patches. This grid-based subdivision (e.g., 5×5, 150x150 px per patch) offers flexibility in patch size and density. The initial formation of these patches can be selected dynamically according to the available resources and the acquisition parameters as illustrated in Figure 2.

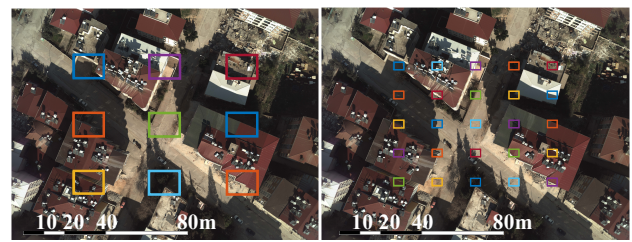

Figure 2: User defined patch illustration

To maintain spatial consistency across frames, the patch translation is guided by GNSS/IMU-assisted transformations in combination with footprint-based surface projection. While GNSS/IMU data from the UAV provide coarse position and orientation information at each time step, we refine the transformation using the projected image footprints, the four-corner projections of the image on the DSM. These quadrilaterals capture the geometric deformation introduced by the terrain and represent the actual geo-referenced surface area covered by the image.

In ideal flat terrain, footprints retain a rectangular shape and correspond well to simple projective transforms. However, as highlighted in Hein et al. (2019), uneven terrain introduces shape distortions in the footprints due to varying elevation (Figure 3 & Figure 4). These deformations invalidate the assumption of a single homography transformation between frames, necessitating local affine approximations or surface-based interpolation to accurately propagate patches.

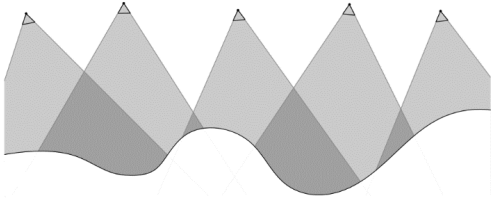

Figure 3: Aerial camera positions and footprints on elevation model Hein et al. (2019)

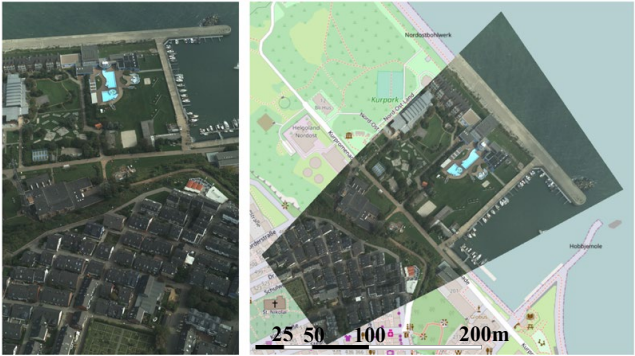

Figure 4: An aerial raw image (left) and its surface-reflected form (right) (Hein et al. (2019))

In the GNSS/IMU-based approach, image-to-image transformation is derived directly from the platform's pose estimates. Each image's position and orientation are obtained from GNSS/IMU readings, and patches are translated based on a rigid-body transform in 3D space, assuming flat terrain or an a-priori known elevation model. The transformation between frames can be computed as:

$$p_{ij}^{(k)} = \pi \left( K * R_k^T \left( \left( T_0 + \lambda * R_0 K^{-1} \overline{p}_{ij}^{(0)} \right) - T_k \right) \right) \quad (3)$$

where
- Camera intrinsic $K$ and UAV poses $(R_k, T_k) \in SO(3) \times R^3$
- $p_{ij}^{(0)} = (u_{ij}, v_{ij})^T$ which is the patch center of the image $I_0$ where $p_{ij}^{(0)} \in R^2$ (similarly, for target image $I_k$)
- $\overline{p}_{ij}^{(0)} = (u_{ij}, v_{ij}, 1)^T$ which are the homogeneous coordinates
- $\pi$ is perspective projection where $\pi(x, y, z) = \left(\frac{x}{z}, \frac{y}{z}\right)$.

- $R_0, R_k$ are camera orientations
- $T_0, T_k$ are camera positions
- $\lambda$ is nominal depth approximation which is usually flat.

In contrast, the second approach we have followed uses surface-consistent footprint projections. Each image is orthoprojected onto the terrain using its camera parameters and DSM comes from TanDEM-X (12m resolution) and SRTM (30m resolution). The four corners of the image define a georeferenced quadrilateral footprint, which inherently encodes the surface deformation. Patches are translated in geographic space according to the local displacement between corresponding footprints across frames. A local planar transformation, in this approach, can be computed per patch as:

$$p_{ij}^{(k)} = R(\theta_k)\left(p_{ij}^{(0)} - c\right) + c + g_k - g_0 \quad (4)$$

where
- Image center location $g_k \in R^2$
- Relative heading $\theta_k \in R$ in plane rotation
- $R(\theta_k) = [cos\theta_k \ sin(-\theta_k); sin\theta_k \ cos\theta_k]$
- $c$ is center of the patch grid
- $g_k$ and $g_0$ are image center positions in global frame (East-North-Up (ENU), ECEF coordinate systems etc.)

As demonstrated in *Equations (3) and (4)*, the footprint-based patch transition method operates independently of camera matrices and depth models, in contrast to the GNSS/IMU-based approach. This independence contributes to a significantly lower computational load. Furthermore, Figure 5 highlights the divergence in patch trajectories between the two methods. While the GNSS/IMU-based method maintains reasonable consistency near the image center, the footprint-based approach exhibits superior performance in preserving spatial alignment across the entire image extent. Therefore, in scenarios where the footprints of all captured images are available in real-time, as is the case in our study, the footprint-based patch transition method not only achieves higher accuracy by incorporating precise terrain information, but also delivers faster processing.

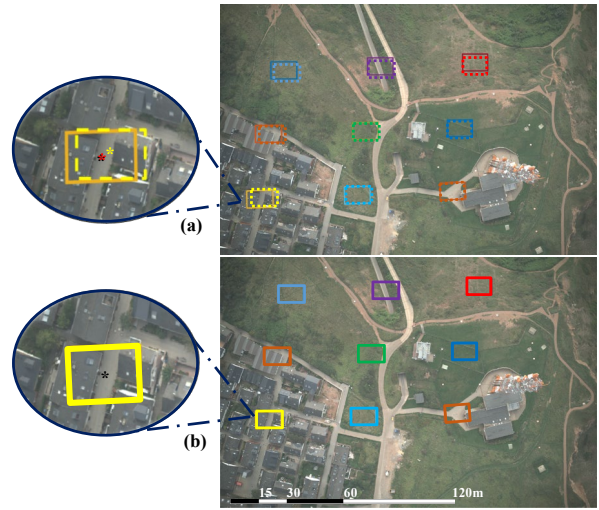

Figure 5: Reflected patches between two consecutive images. Dashed frames: GNSS/IMU method; solid frames: footprint method. In (a), red and yellow asterisks denote the centers of the solid and dashed frames, respectively. The black asterisk in (a) and (b) marks the same location for reference.

After translation, if a patch moves out of the current image bounds, it is re-initialized at the opposite edge to achieve continuous flow across the image stream. This process allows persistent tracking of features over large temporal windows without loss of coverage.

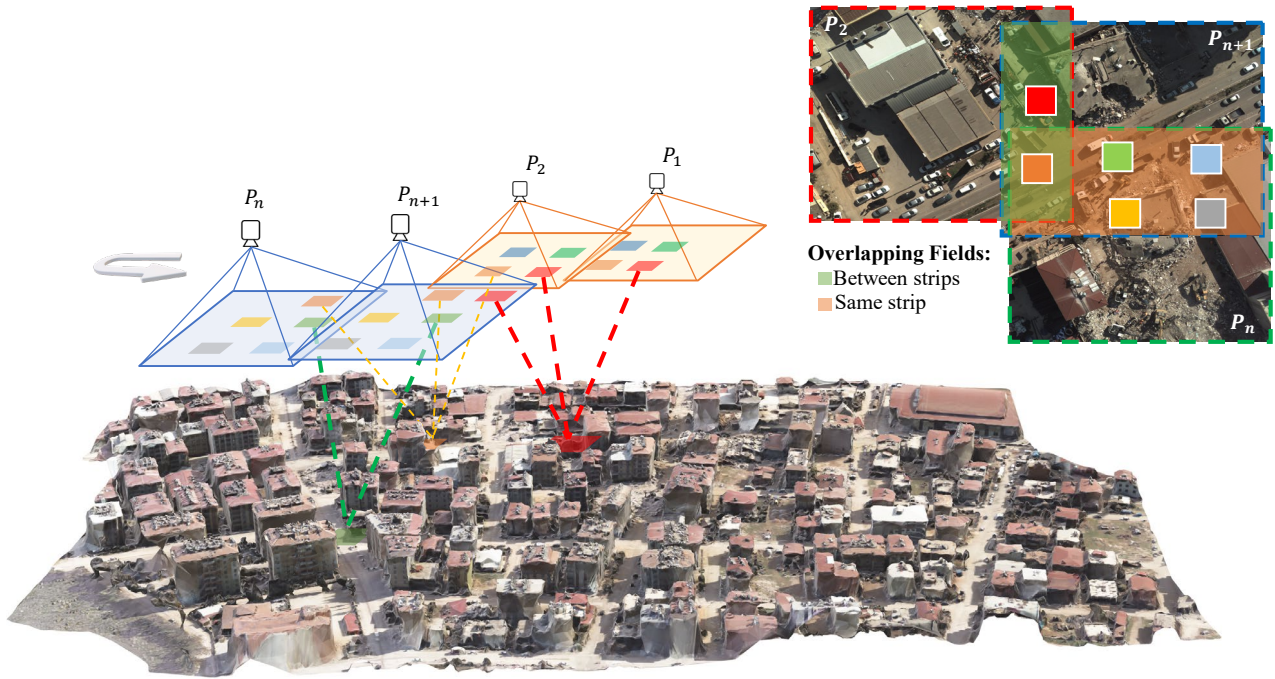

Figure 6: Patches in cross-strips

When transitioning between neighbouring flight strips, overlap between newly acquired images and previously captured adjacent-strip images is detected by comparing their projected footprints. The previously defined patches in overlapping regions are then re-projected onto the current image, enabling cross-strip feature tracking. This ensures that orientation and alignment are maintained not just within but also between strips, a major limitation in many conventional approaches.

Finally, features extracted within each translated patch are matched with their corresponding patch IDs across frames (Figure 6). This patch-based feature matching strategy significantly reduces the computational load by narrowing the search space to predefined, locally consistent regions, making the high-resolution matching process feasible for real-time onboard execution. These matched features are then passed to the continuous bundle adjustment module to refine pose estimates and maintain global consistency.

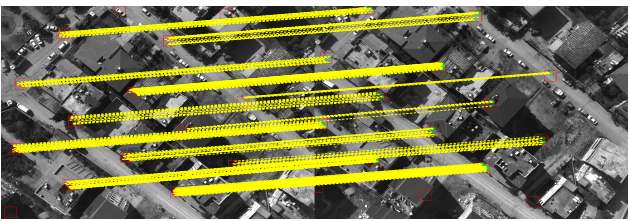

Figure 7: Feature matching in corresponding boxes

**3.2. Localized Bundle Adjustment for Real-Time Processing**
Real-time processing of 50 MP+ images exceeds the capabilities of traditional BA methods. While global BA offers high accuracy, it is computationally prohibitive for large datasets. Incremental BA is faster but relies on tracks established relative to only one previous image which makes it prone to drift and lacks inter-strip consistency. To overcome these issues, we introduce a cluster-based BA that processes user-defined image groups, integrating overlapping frames from neighbouring strips to ensure smooth transitions and global coherence.

In the proposed method, the user defines a cluster size $M$ (i.e., the number of images to be optimized simultaneously). Within each cluster, feature tracks are established based on matched features in corresponding patches. Initially, each image is assigned an approximate pose from GNSS/IMU data, with the position and orientation provided in the ECEF/ENU coordinate systems. These initial estimates, denoted as $T_k^{(0)}$ and $R_k^{(0)}$ for $I_k$, serve as a starting point for the BA optimization.

The reprojection error for a given observation of a 3D scene point $X_j$ in image $I_k$ is given by

$$p_{kj} = \pi\left(KR_k^T(X_j - T_k)\right) \quad (5)$$

where $\pi, K$ and $I_k$ represent same annotations as in the previous section. The corresponding BA cost function for a set of images $I_k$ and feature tracks is formulated as

$$E_{BA} = \sum_{k \in I} \sum_{j \in J_k} \rho\left(\left\|p_{kj} - \pi\left(KR_k^T(X_j - T_k)\right)\right\|^2\right) \quad (6)$$

with $\rho$ being a robust cost function to mitigate the influence of outliers.

Global BA seeks to solve the above minimization over the entire dataset, optimizing all camera poses $\{P_k = (R_k, T_k)\}$ and 3D points $\{X_j\}$ simultaneously. Traditional incremental bundle adjustment lacks a cross-check mechanism and relies solely on consecutive images. While this reduces computational load, it can lead to drift and fails to incorporate overlaps between adjacent flight strips. Furthermore, when not confined to a local cluster, the track list grows continuously with each iteration, resulting in a significant slowdown.

In contrast, our cluster-based BA approach operates on a dynamically formed cluster $C_l$ defined as

$$C_l = \{I_{k_1}, I_{k_2}, \dots, I_{k_M}\}$$

where $M$ is the selected trade-off between computational load and local reconstruction accuracy. The cost function within each cluster is

$$E_{C_l} = \sum_{k \in C_l} \sum_{j \in J_k} \rho\left(\left\|p_{kj} - \pi\left(KR_k^T(X_j - T_k)\right)\right\|^2\right) \quad (7)$$

By using the georeferenced ECEF coordinates as the initial positions, the optimization directly yields geo-referenced camera poses. Feature tracks are established within the cluster by matching corresponding patches in real time; concurrently, the algorithm verifies the existence of overlapping images from neighbouring flight strips (Figure 7). When the number of images in the current cluster reaches the predefined threshold, the BA optimization is executed for that cluster.

To ensure smooth transitions between successive clusters, we enforce an overlap between clusters. Specifically, the last 25% of the images in the previous cluster $C_{l-1}$ are used as references for the new cluster $C_l$. Within this overlapping set, the first half of images are fixed (i.e., their optimized poses from the previous cluster are retained), while the orientations of the remaining images are adjusted using a weighted averaging scheme based on the number of matched features. For a common overlapping image $I_o$, its updated pose is computed as

$$P_0^{new} = \frac{\omega_0^{(old)} P_0^{C_{l-1}} + \omega_0^{(new)} P_0^{C_l}}{\omega_0^{(old)} + \omega_0^{(new)}} \quad (8)$$

where $\omega_0^{(old)}$ and $\omega_0^{(new)}$ represent the weights proportional to the number of feature matches in the old and new clusters, respectively. This scheme ensures that the transition between clusters is smooth and consistent, thereby linking the adjusted orientations and positions across different clusters.

### 3.3. Validation and Benchmarking

To evaluate the performance of the proposed multi-strip cluster-based bundle adjustment, we compare it against two established methods: (1) Traditional incremental BA, which is more suitable for real-time applications than global BA; and (2) Cluster-based (local) incremental BA, which run in different cluster of images similar to proposed method, and resets the track list in each new cluster entrance, which is one of the main reason of performance deceleration of incremental BA while the number of images increase.

We employ the following evaluation metrics to assess both geometric accuracy and computational efficiency:
- Qualitative accuracy of oriented images: to quantify alignment precision and global consistency.
- Reprojection errors (mean and standard deviation) of the reconstructed point clouds: to measure the geometric fidelity of the estimated scene.
- Execution time per image pair (measured in MATLAB): to highlight the real-time suitability and relative performance gains of the proposed method.
- Feature extraction and feature matching time: to evaluate the proposed patch-based matcher performance.

## 4. Results and Discussion

The proposed method was evaluated using a 60-image, two-strip dataset captured with the DLR MACS during an emergency mapping mission following the 2023 Turkiye Earthquake. Each image has a resolution of 7920×6004 pixels. The tests were conducted in MATLAB without GPU support, using a standard office laptop equipped with a 13th Gen Intel Core i7 processor (20-core) and 32 GB RAM.

Two established bundle adjustment (BA) strategies were used for comparison. The first is a conventional incremental BA approach, commonly adopted in SLAM pipelines, which sequentially incorporates new images. The second is a cluster-based incremental BA strategy, where images are processed in local clusters to prevent excessive memory growth due to extended feature tracks. Each cluster contains 12 images, with the last three images reused as references in the subsequent cluster, similar to the proposed method. Both baseline methods operate on the full-size images without any downsampling.

The proposed method employs a 4×4 patch grid per image, with each patch measuring 100×100 pixels. These patches are tracked using GNSS/IMU data and DSM-informed footprint transformations.

In terms of geometric accuracy, the proposed method achieves comparable performance to the incremental BA. The mean reprojection error decreased from 0.727 px (incremental BA) to 0.710 px, and the standard deviation reduced from 1.756 px to 1.075 px. Though the numerical differences are modest, they indicate improved consistency across image strips. This stability is attributed to localized BA clusters with overlap-aware initialization based on projected image footprints (Table 1).

Runtime performance exhibits more pronounced differences. The total processing time dropped from 962.69 seconds (incremental BA) to 66.45 seconds with the proposed method. Feature matching time alone was reduced from 530.31 seconds to just 0.62 seconds, thanks to patch-constrained matching that avoids brute-force comparisons. Feature extraction time also improved, decreasing from 61.27 to 48.86 seconds by limiting detection to patch windows, which suppresses irrelevant features and reduces false correspondences (Table 1).

The cluster-based incremental BA achieved a moderate runtime of 14.73 seconds per image pair and showed better memory efficiency compared to the incremental BA. However, qualitative analysis in Figure 8 reveals that its orientation estimates degrade when transitioning between strips. While intra-strip consistency is preserved, inter-strip transitions show drift due to the absence of inter-cluster correspondence management (Figure 8a).

Table 1: Comparison of incremental BA and the proposed method

|  | Incremental BA | The Proposed Method |
|---|---|---|
| Mean of Reprojection Error (px) | 0.72743 | **0.71049** |
| Std-dev of Reprojection Errors (px) | 1.7563 | **1.0758** |
| Total Feature Extraction Time (sec) | 61.27 | **48.86** |
| Total Feature Matching Time (sec) | 530.31 | **0.62** |
| Total Run Time (sec) | 962.69 | **66.45** |
| Run Time per Image Pairs (sec) | 16.04 | **1.11** |

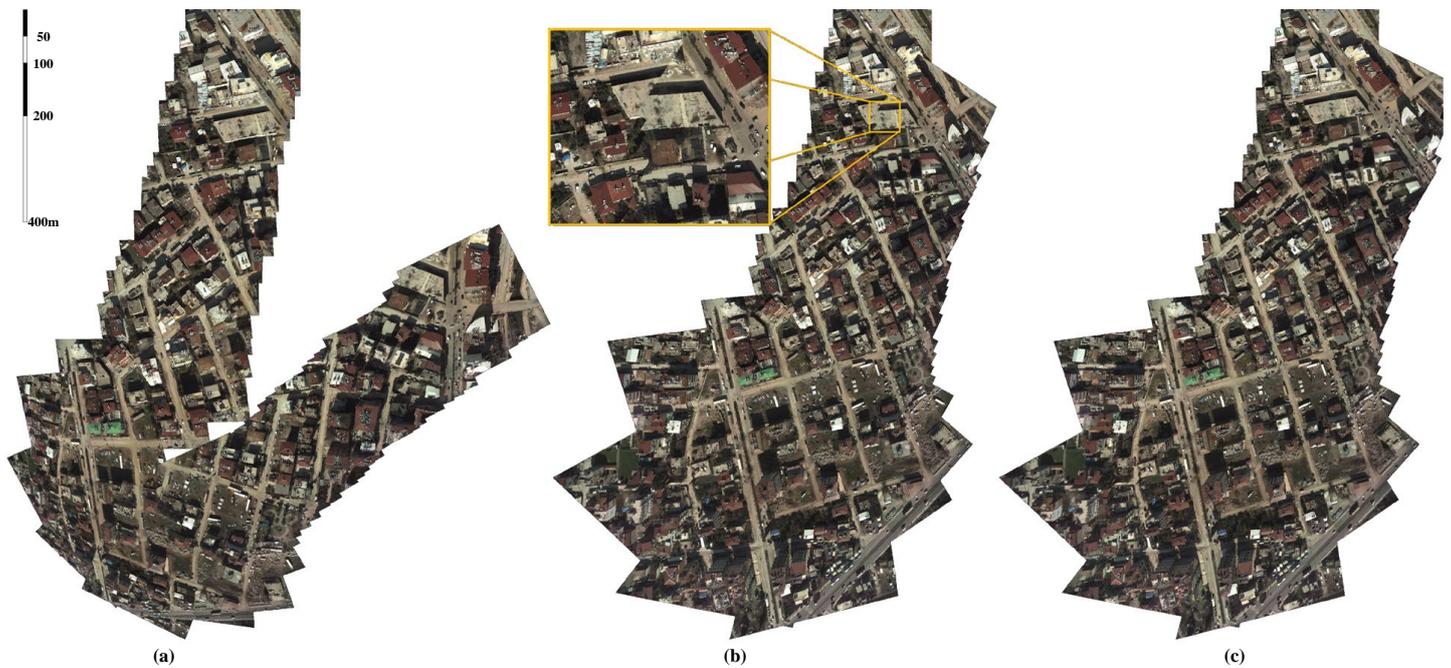

Figure 6: Comparison of cluster-based incremental BA (a), incremental BA (b), the proposed method (c)

In contrast, the incremental BA result (Figure 8b) suffers from accumulated positioning error, manifesting as parallax distortions, a known issue in long-track incremental pipelines. The proposed method effectively avoids such artifacts by maintaining inter-strip consistency through footprint-guided neighbor strip identification and localized clustering. Figure 8c illustrates that the proposed method produces geometrically consistent results across strips without the parallax or drift issues observed in the other approaches.

## 5. Conclusion

This study presents a practical and efficient framework for real-time bundle adjustment of high-resolution UAV imagery, addressing the limitations of conventional approaches that are either too slow or require image downsampling. By dividing full-resolution images into user-defined patches and leveraging UAV GNSS/IMU data alongside terrain information, the method enables accurate feature tracking and localized bundle adjustment within manageable clusters. The inclusion of overlapping images from neighbouring strips ensures spatial consistency and mitigates drift commonly observed in incremental methods.

In the context of the research question, whether full-resolution UAV imagery can be processed in real-time without compromising orientation accuracy, the results are encouraging. The method demonstrates that, through patch-based tracking and cluster-wise optimization informed by geo-referenced data, it is possible to achieve a practical trade-off between computational speed and geometric fidelity. This makes it feasible to deploy the method onboard UAVs or in edge-computing environments, especially in scenarios where fast decision-making is critical.

While the method offers flexibility through user-defined hyperparameters such as patch size and the number of images per cluster, this reliance on manual tuning can be time-consuming, especially for new users. Future work should focus on automating the selection of these parameters based on image resolution and mission scale to improve usability and repeatability. Additionally, to broaden accessibility, a variant of the algorithm should be developed that relies solely on GNSS/IMU data when image footprints are unavailable.

Overall, the proposed method bridges the gap between high-fidelity mapping and real-time processing, making it well-suited for time-critical applications such as disaster response, where both accuracy and speed are imperative.